\documentclass[letterpaper, 10 pt, conference]{ieeeconf}  

\IEEEoverridecommandlockouts                              

\usepackage{graphics} 
\usepackage{epsfig} 
\usepackage{mathptmx} 
\usepackage{times} 
\usepackage{multirow}
\usepackage{amsmath} 
\usepackage{amssymb}  
\usepackage{algorithmic}
\usepackage{algorithm}
\usepackage{balance}
\usepackage{soul,xcolor}
\usepackage[hidelinks, implicit=false]{hyperref}
\let\labelindent\relax
\usepackage{enumitem}
\usepackage{scalerel}
\setitemize{leftmargin=*}

\title{\LARGE \bf
Generating a Terrain-Robustness Benchmark for Legged Locomotion: A Prototype via Terrain Authoring and Active Learning
}

\author{Chong Zhang$^{1\dagger}$ and Lizhi Yang$^{2}$
\thanks{$^{\dagger}$Corresponding author. Email: {\tt\small chozhang@ethz.ch}}%
\thanks{$^{1}$Chong Zhang is with the Department of Mechanical and Process Engineering, ETH Zurich, Switzerland. Email: {\tt\small chozhang@ethz.ch}}%
\thanks{$^{2}$Lizhi Yang is with the Department of Mechanical and Civil Engineering, California Institute of Technology, United States. Email: {\tt\small lzyang@caltech.edu}}%
}

\begin{document}

\maketitle
\thispagestyle{empty}
\pagestyle{empty}
\begin{abstract}

Terrain-aware locomotion has become an emerging topic in legged robotics. However, it is hard to generate diverse, challenging, and realistic unstructured terrains in simulation, which limits the way researchers evaluate their locomotion policies. In this paper, we prototype the generation of a terrain dataset via terrain authoring and active learning, and the learned samplers can stably generate diverse high-quality terrains. We expect the generated dataset to make a terrain-robustness benchmark for legged locomotion. The dataset, the code implementation, and some policy evaluations are released at https://bit.ly/3bn4j7f.

\end{abstract}

\section{Introduction}

Terrain-aware locomotion has been gaining prominence in the domain of legged robotics \cite{fankhauser2018robust} \cite{tsounis2020deepgait} \cite{miki2022learning} \cite{gangapurwala2022rloc}. To overcome all kinds of unstructured terrains in the wild, legged robots need great robustness in locomotion. However, such robustness is difficult to quantify, especially for the unstructured terrains that can hardly be foreseen. As a result, existing works tend to manually build specific terrains and report the success rates for several toy experiments \cite{fankhauser2018robust} \cite{gangapurwala2022rloc} \cite{jenelten2020perceptive} \cite{Jenelten2022TAMOLS}.

\begin{figure}[t]
  \centering
    \includegraphics[width=83mm]{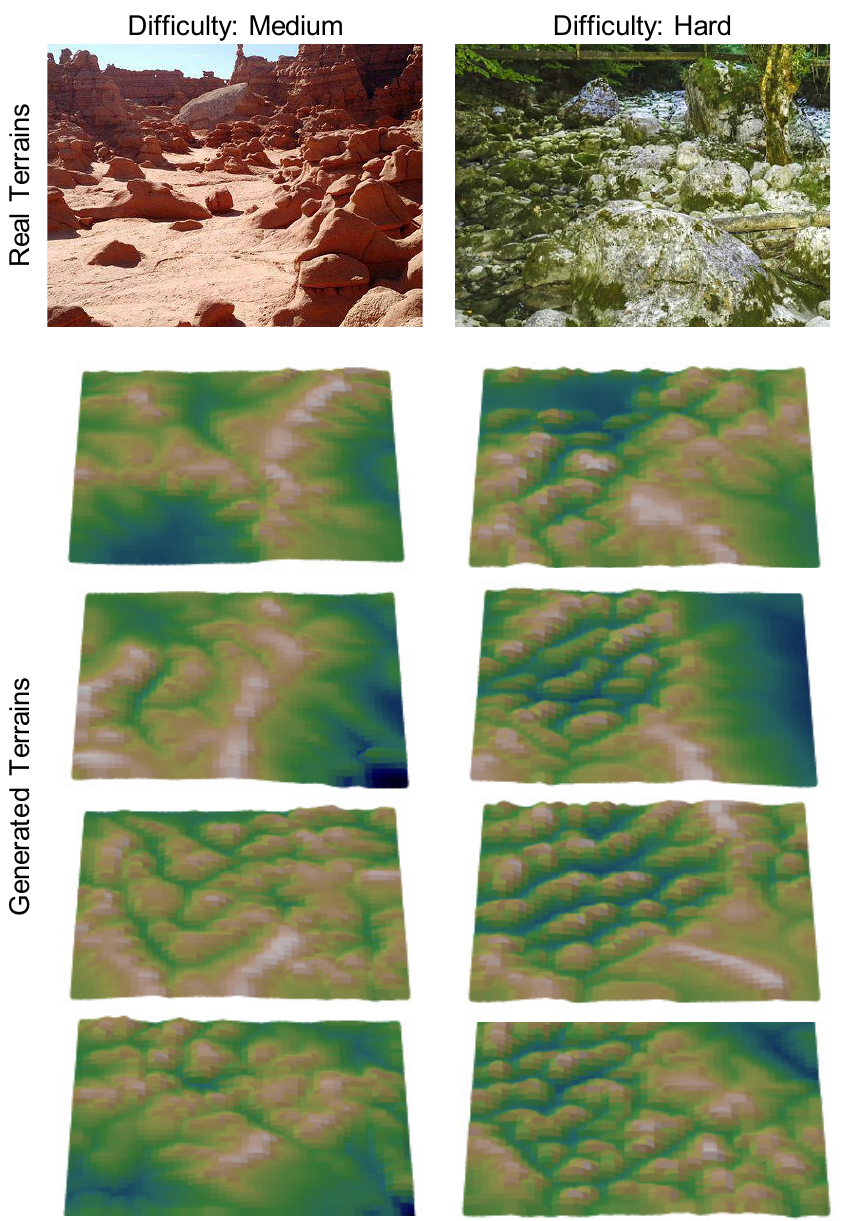}
    \caption{
       Cherry-picked photos of real terrains and visualization of generated terrains in the benchmark. The first two rows are pairs where the generated terrain samples resemble the real terrains. The real terrains are irrelevant to dataset generation, but reflect how the generated terrains are realistic. The other three rows compare the terrains generated for two difficulty levels, "medium" (left) and "hard" (right), where the pairs in the same row share somewhat similar terrains but the "hard" one is more complex and contains more ravines.
    } 
    \label{fig:sample_demo}
  \vspace{-5mm}
\end{figure}

In light of this, our paper seeks a way to statistically quantify the robustness in terms of terrains for different locomotion policies by building a terrain-robustness benchmark for legged locomotion. The terrain samples should be diverse and resemble those in the wild, so that the robustness can be measured and improved in simulation. Furthermore, these terrains should be challenging to traverse in order to discriminate between different policies. Different from \cite{Jenelten2022TAMOLS} \cite{lee2020learning}, terrains generated in this paper are more unstructured rather than parameterized. Results also show that, unlike structured terrains where people can easily assign parameters, generation of high-quality unstructured terrains is difficult. Fig. \ref{fig:sample_demo} exhibits some of our terrain samples in the benchmark dataset. To our knowledge, no existing work has built such a benchmark, and this paper makes a prototype.

To this end, three challenges must be solved: 1) to achieve reliable quantification of robustness, the terrain samples should resemble real terrains in the wild; 2) to achieve the easy generation of high-quality terrains (i.e., be challenging to a user-specified extent), the generation process should be somewhat controllable; 3) the sampler must maintain terrain quality and diversity simultaneously. Among these challenges, the first two can be viewed as mapping a controlled input to a realistic terrain, and the last challenge is to obtain a sampling policy that can generate diverse high-quality inputs. Hence the problem is naturally divided into two parts: one for terrain generation, and one for sampling.

With the concept of fractal \cite{polidori1991description} \cite{shaker2016fractals}, we believe that a local terrain elevation map can be obtained via resizing a realistic terrain of any appropriate scale, including the scales of the terrain data obtained by remote sensing techniques. Based on this and referring to \cite{guerin2017interactive}, we use the conditional generative adversarial networks (GANs) to generate terrains from specified control points as the inputs, and the training dataset that the outputs should resemble comes from the US map consisting of digital elevation models \cite{gesch2002national}. In this way, together with some tricks, we can generate realistic and unstructured terrains from small-size control point sets, which tackles the first two challenges.  

To tackle the third challenge, we model the generation of control points as a Markov decision process (MDP) on a directed acyclic graph (DAG), and refer to the arising active learning method generative flow networks (GFlowNets) \cite{DBLP:journals/corr/abs-2111-09266} to get the control point samplers. It has been shown that, such an active learning method not only outperforms sampling-based methods in data efficiency, but also maintains a high diversity of candidates compared with reinforcement learning methods \cite{bengio2021flow}. In this paper, the learned samplers can generate diverse hard-to-obtain samples or sets of control points, with high quality regarding legged locomotion.

\begin{figure}[t]
  \centering
    \includegraphics[width=87mm]{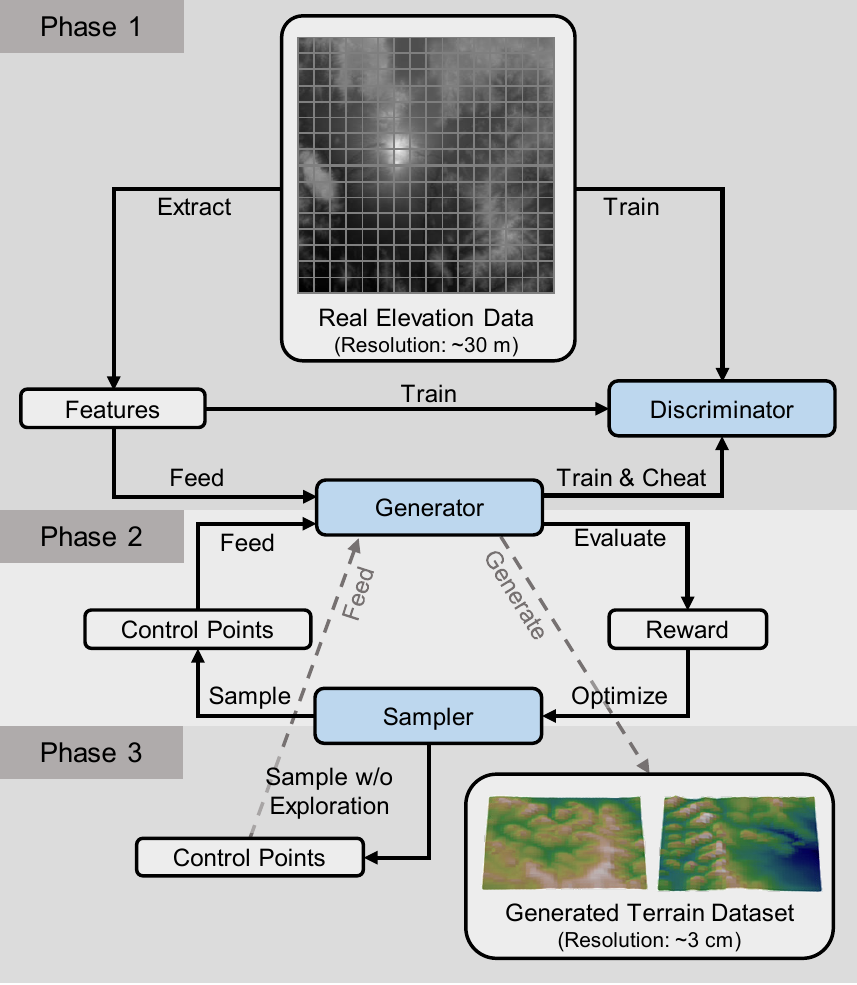}
    \caption{
       The pipeline of the proposed dataset generation method. In phase 1, a generator and a discriminator are trained to ensure that the generated terrains are realistic. In phase 2, the sampler learns through exploration and exploitation to sample high-quality control points for terrain generation. In phase 3, the trained sampler does no more exploration and is used to sample high-quality terrains for the dataset.
    } 
    \label{fig:pipeline}
  \vspace{-5mm}
\end{figure}

Based on how the three challenges are met, this paper proposes a pipeline to learn the desired sampler, as is shown in Fig. \ref{fig:pipeline}. Three models are trained: 1) a terrain generator, 2) a terrain discriminator, and 3) a control-point sampler. The terrain discriminator ensures that the generated terrains are realistic. Using the sampler and the terrain generator, we can sample diverse high-quality terrain samples after only a few iterations, without costly real data collection (e.g., in \cite{collect_terrain}).

Our contributions in this paper are summarized as follows:
\begin{enumerate}
    \item An adaptation of realistic terrain authoring to take small-size sets of control points as the inputs;
    \item Learned control point samplers to generate high-quality terrain samples for legged locomotion;
    \item The generation of the first benchmark dataset to our knowledge to statistically evaluate the robustness of legged locomotion on unstructured terrains.
\end{enumerate}

\section{Related Works}

\subsection{Terrain-aware legged locomotion}

This paper tries to generate a terrain-robustness benchmark for terrain-aware legged locomotion, which is an emerging topic in the robotics community. Generally speaking, there are two kinds of solutions to terrain-aware locomotion: optimization-based methods and learning-based methods. 

Optimization-based methods typically obtain reasonable footholds and trajectories through searching and optimization, with heuristic scores or losses \cite{jenelten2020perceptive} \cite{Jenelten2022TAMOLS}. Often, the robustness tests are done on stairs or stacked bricks and tiles, without many trials or high terrain diversity.

Learning-based methods learn trajectories \cite{gangapurwala2021real}, footholds \cite{tsounis2020deepgait} \cite{gangapurwala2022rloc} \cite{iit2019ral} or joint-level outputs \cite{miki2022learning} \cite{rudin2022learning} \cite{acero2022learning} in a data-driven way. Although various terrains are generated in simulation to feed the data-hungry learning process, they either consist of simple stairs and steps, or are non-realistic and generated from uncontrollable noises. Despite some successful deployments in the wild \cite{miki2022learning}, robustness tests are done either on limited real terrains, or statistically on non-realistic low-quality ones in simulation, on top of the heuristically defined structured terrains.

\subsection{Terrain generation}

Terrains, in the domain of terrain-aware legged locomotion, are typically represented by heightmaps, i.e., two-dimensional matrices of real numbers indicating the height at different points. A traditional method for terrain generation is to use Perlin noise \cite{lagae2010survey}, as is adopted by existing works \cite{miki2022learning} \cite{lee2020learning}. Although policies can be trained in simulation with such terrains, verifications must be done on real robots after sim-to-real transfer, because using Perlin noise does not lead to realistic heightmaps \cite{shaker2016fractals}.

Alternative methods are to generate fractal terrains, e.g., to use the diamond square algorithm \cite{fournier1982computer} and the fractal brownian motion algorithm \cite{mandelbrot1968fractional}. However, it is difficult to regard them as realistic. 

An emerging way to generate realistic terrains is to use GANs \cite{goodfellow2014generative}, where a discriminator tries to classify whether a sample comes from the dataset, and a generator tries to cheat the discriminator by generating samples from noises. Examples of GAN-based terrain generation are \cite{wulff2017deep} and \cite{spick2019procedural}. Yet, to achieve partially controllable generation and actively generate a dataset, we need interactive terrain authoring based on conditional GANs \cite{guerin2017interactive}. To be specific, the discriminator classifies whether the samples together with certain features are from the training dataset, and the generator generates fake samples from not only noises but also the features. Finally, the generator can generate realistic terrains from given input features, and the noises only affect small-scale details. 

\subsection{Active learning}

Active learning is a technique that selects a subset of all possible candidates and gets them labeled to improve the model performance \cite{settles2009active}. In this paper, we adopt this technique as it can be extremely hard and data-inefficient to directly get high-quality terrains by random sampling. Also, it can be time-consuming to evaluate a terrain, e.g., reporting the performance of multiple gaits on the same terrain with varying dynamical parameters, although we would only use a heuristic terrain score for a prototype in this paper.

Unlike the common way active learning is applied to get more labeled data, in this paper, we aim to achieve an effective sampler that can provide a diverse set of high-quality terrains. The recently proposed GFlowNets \cite{DBLP:journals/corr/abs-2111-09266} has provided a promising solution where candidates are sampled in proportion to their given rewards. Different from reinforcement learning methods that only provide a low-variance policy to maximize the expected return \cite{sutton2018reinforcement}, GFlowNets can maintain the diversity of selected candidates, which has been empirically verified in domains such as molecule synthesis \cite{bengio2021flow}. Another choice for diverse sampling is the iterative Markov chain Monte Carlo (MCMC) methods \cite{van2018simple} \cite{grathwohl2021oops} which are data-inefficient and vulnerable to local exploration.

\section{Generating Terrains as Trajectories}
\label{sec:generating}
In this section, we present how we generate realistic terrains from features or control points, and how the terrain generation can be viewed as an MDP.

\subsection{Conditional GANs for terrain authoring}

\begin{figure}[t]
  \centering
    \includegraphics[width=86mm]{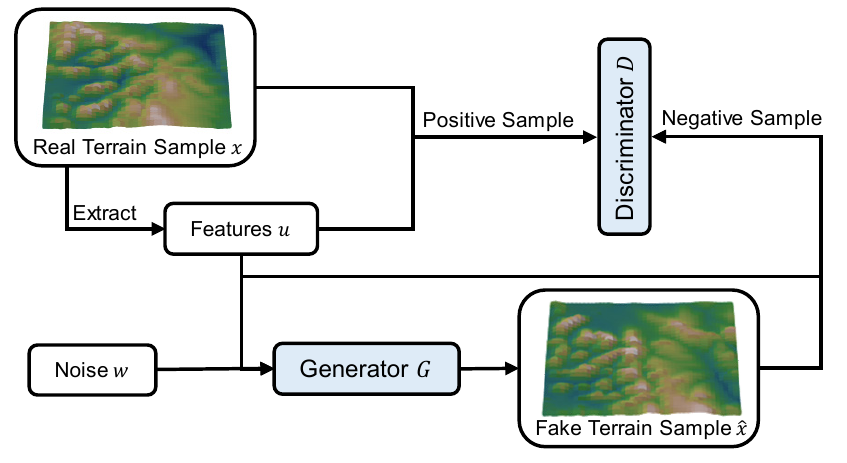}
    \caption{
       The conditional GAN system for terrain authoring. The discriminator ${D}$ tries to distinguish the positive samples and the negative samples, while the generator ${G}$ learns to fool the discriminator ${D}$.
    } 
    \label{fig:gan}
  \vspace{-4mm}
\end{figure}

As is mentioned, we use the interactive terrain authoring method in \cite{guerin2017interactive} to generate terrains from controlled inputs, which is based on conditional GANs. The training of the model is illustrated in Fig. \ref{fig:gan}. For each sample $x$ in the dataset, we extract features $u$ from it, and $(x,u)$ makes a positive sample for the discriminator ${D}$. With the generator ${G}$ and the random noise $w\sim\mathcal{N}(\mathbf{0},\mathbf{I})$, ${G}(u,w)$ denotes the generated fake sample, and $({G}(u,w),u)$ makes a negative sample for ${D}$. With the positive sample and the negative sample, ${D}$ predicts the pixel-level classification for $x$ and is trained with the binary crossentropy loss:
\begin{equation}
\begin{split}
     L_{D}(x,u,{G},w) = &-\mathbb{E}_{\rm pixel}[\log({D}(x,u))\\
     &+\log(1-{D}({G}(u,w),u))].   
\end{split}
\end{equation}
Here we take the pixel-level prediction to ensure detailed terrain outputs according to \cite{isola2017image} and a third-party implementation \cite{sketch-to-terrain} of \cite{guerin2017interactive}.

The generator ${G}$ learns to fool the discriminator ${D}$. Following \cite{isola2017image} and \cite{sketch-to-terrain}, the loss for ${G}$ is defined as:
\begin{equation}
\begin{split}
        L_{G}(x,u,{D},w) =  \mathbb{E}_{\rm pixel}[-\lambda^G_1\log({D}({G}(u,w),u))\\
        +\lambda^G_2|x-{G}(u,w)| ],
\end{split}
\end{equation}
where $\lambda^G_1$ and $\lambda^G_2$ are manually defined coefficients, and $|\cdot|$ denotes the absolute error.

\subsection{Implementation details}
\subsubsection{Dataset}
Regarding real heightmaps, we cherry-picked some images with rich elevation information from the USGS 3D elevation dataset \cite{gesch2002national}, which can be reproduced in our code. The images are of 1 arc-second resolution ($\approx 30\ \rm m$), and are of size $(3600,3600)$. We split each image into $256$ patch samples of size $(225,225)$. 

\subsubsection{Target resolution}
Our initial purpose is to generate $2.5\rm\ cm$-resolution heightmaps, leading to a total size of $\sim 5.6 \rm\ m \times 5.6\ m$, and the maximum height is $10\times 2.5\rm\ cm$ if no additional slope added. This is specified for the size and mobility of some popular legged robots, e.g., ANYmal \cite{hutter2016anymal} and A1 \cite{a1}. Despite the different resolution from the dataset, with the idea of fractal, we believe the used elevation data can be rescaled to the size we want while remaining realistic, as is verified in Fig. \ref{fig:sample_demo}. The generated terrains can also be rescaled to other resolutions, e.g., $3\rm\ cm$ or $5\rm\ cm$.

\subsubsection{Selected features and masking}
In \cite{guerin2017interactive}, different kinds of features are extracted: rivers, ridges, peak points, and basin points. However, it is intractable to automatically draw rivers and ridges as humans intuitively do. Also, we found basin points sometimes hard to detect in our dataset. Hence, we detect pits instead via the pysheds library \cite{bartos_2020}, and detect peaks by inverting the elevation for pit detection. The features we used are peak points and pit points, and they are represented as 0-1 matrices of size $(225,225)$ to feed the models. 

Yet, in this way, there can be too many features on one terrain. Sometimes, up to 200 feature points can be detected, which makes it hard to generate high-quality sets of control points. Also, we found that feature points are often too intensive at certain areas. Thus, borrowing insights from masked autoencoders \cite{he2022masked}, we believe most of the points can be redundant and we randomly masked the points with a probability of 75\% per visit. In brief, extracted features in our paper are peak points and pit points that are masked with a probability of 75\%.

\subsubsection{Models}
We refer to the model structures in \cite{sketch-to-terrain}, where the generator is a U-Net \cite{ronneberger2015u} as in \cite{isola2017image}, and the discriminator consists of several convolutional layers, finally activated by the sigmoid function for pixel-level prediction. Despite the noise inputs, the output terrains only vary little in small-scale details for fixed features after training, which can be omitted.   

\subsubsection{Low-pass filtering}
After training, the generator generates realistic terrains as expected. However, due to the instability of neural network outputs, there can be spikes. Thus, we apply a low-pass filter to the generated terrains as post-processing. To be specific, after empirical evaluation, we apply Gaussian blurring with kernel size $(5,5)$ and $\sigma_x=\sigma_y=1$ for unscaled height outputs in $[-1,1]$. The heights are then rescaled to $[0,\text{maximum height}]$.

\subsection{A DAG for terrain generation}
The terrain generation can be viewed as MDP trajectory generation, where a set of control points as the input features mostly determines the generated terrain. Specifically, starting from an empty control point set, the MDP goes as follows:
\begin{enumerate}
    \item Three kinds of actions can be taken: to terminate, to add a peak point, or to add a pit point;
    \item If the action is to terminate, a terrain is generated with the existing control points, otherwise the new point is added to the control point set.
\end{enumerate}

To make the problem tractable, we assume:
\begin{enumerate}
    \item Two control points that are too close to each other do not make sense, and the small change in the point positions does not lead to significant changes in the terrain. 
    \item There cannot be too many control points, which will only lead to redundancy and complexity.
\end{enumerate}
Thus we discretize the rows and the columns to $75\times 75$ grids for actions, which means that one grid represents the center of $3\times 3$ pixels. Each point-adding action occupies a grid, limiting the number of feasible actions to $2\times 75^2+1$ at most. We also force the MDP to terminate after there have been $60$ points, which limits the complexity of the problem.

The states, as assumed above, can be defined as a set of points:
\begin{equation}
    s=\{p_1,\dots,p_l\}, 0\le l \le 60,
\end{equation}
where $s$ is empty if $l=0$. Each point is the combination of its position and its attribute (peak or pit). Mathematically, we represent each point as a 3-d vector, where the first dimension is the $x$ coordinate normalized to $[-1,1]$, the second dimension is the $y$ coordinate normalized to $[-1,1]$, and the third dimension is $1$ for peaks and $-1$ for pits. Thus the features $u=u(s)$ are easily determined by the state $s$.

The action $a$ is also represented as a 3-d vector except for the termination. It is the 3-d vector to represent the added point in the state.

Such an MDP makes a DAG because, one state can have multiple (i.e. the number of points in the state) parent states, and a fewer-point state cannot have a more-point parent state. Denoting the transition of states via actions by $s'=T(s,a)$, we have:
\begin{equation}
    \mathrm{card}(T(s,a)) = \mathrm{card}(s)+1,
\end{equation}
where $\mathrm{card}(\cdot)$ denotes the cardinality of a state. A trajectory $\tau$ can be represented as $\tau=(s_0,\dots,s_n)$ in that the transition is deterministic, and $s_0$ is exactly $\emptyset$. The trajectory length is defined as the cardinality $n$ of the final state $s_n$.

\section{GFlowNets for Terrain Generation}
In this section, we detail our method to generate high-quality control point sets via GFlowNets.

\subsection{Formulation and optimization}
Equivalent to but slightly different from GFlowNets in \cite{DBLP:journals/corr/abs-2111-09266} \cite{bengio2021flow}, in our configuration for the state space $\mathbb{S}$ and the action space $\mathbb{A}$, we define an ideal flow function $F:\mathbb{S}\times \mathbb{A} \rightarrow \mathbb{R}^+ $ s.t.
\begin{equation}
    \label{eq:flow}
    \sum_{T(s,a)=s'}F(s,a) = \sum_{a \in \mathbb{A}^*(s') }F(s',a), \forall s' \in \mathbb{S}\backslash\{\emptyset\},
\end{equation}
where $\mathbb{A}^*(s')$ denotes the feasible action set of $s'$, i.e., to terminate the MDP or to add a peak or pit point at an unoccupied grid. For the termination action $a^T$, the following boundary conditions are satisfied:
\begin{equation}
    \label{eq:boundary}
    F(s,a^T) =R_G(s)= \mathbb{E}_{w} R\left(G\left(u\left(s\right),w\right)\right), \forall s \in \mathbb{S},
\end{equation}
where $R(\cdot)$ is the reward for a generated terrain. Because the reward for the same $s$ varies little w.r.t. the noise $w$, we can sample only once to estimate the termination flow. Finally, when the sampling probability goes as 
\begin{equation}
    P(a|s) = \frac{F(s,a)}{\sum_{a'\in \mathbb{A}^*(s)}F(s,a')}\propto F(s,a), \forall a \in \mathbb{A}^*(s),
\end{equation}
the probability of sampling a final state $s^T$ at termination satisfies
\begin{equation}
\label{eq:property}
    P(s^T) = \frac{R_G(s^T)}{\sum_{s\in \mathbb{S}} R_G(s)} \propto R_G(s^T), \forall s^T \in \mathbb{S}.
\end{equation}
We refer readers to \cite{DBLP:journals/corr/abs-2111-09266} and \cite{bengio2021flow} for the proof.

Yet, since $s$ is of size $60$ at maximum and $A^*(s)$ can contain $>10000$ feasible actions, we have to approximate the flows with a neural network. In this paper, we approximate the logarithm $\hat{F}_{\log}$ of the flows, and the flow matching loss w.r.t. each trajectory $\tau$ is defined as follows for optimization:

\begin{equation}
\begin{split}
    \label{eq:optim}
    L_F&(\tau)=\sum_{s'\in \tau \backslash \{s_0\}}\left(
    \log\left[\epsilon+\sum_{T(s,a)=s'}\exp \hat{F}_{\log}(s,a)\right]
    \right.\\
    &\left. -  \log\left[\epsilon+ R_G(s') +
    \sum_{a \in \mathbb{A}^*(s')\backslash \{a^T\} } \exp \hat{F}_{\log}(s',a) \right]
    \right)^2,
\end{split}
\end{equation}
where $\epsilon$ is a small positive value and we take $10^{-6}$. During training, new trajectories are sampled with approximated flows according to (\ref{eq:property}), and the approximation model is optimized according to (\ref{eq:optim}) on a batch of trajectories in an "off-policy" fashion.

\subsection{Scoring and rewarding}
The generated terrains are to be evaluated to select the challenging ones for robots to traverse. This can be done from many aspects, e.g., to test different policies on the terrain. In this paper, we use the heuristic foothold score as in \cite{jenelten2020perceptive} to prototype the learning-based dataset generation. Such a score was used to represent the safety of a foot location, and we use the mean score to indicate the difficulty over the whole terrain. To be specific, the score $\beta$ for a terrain heightmap $h$ is defined as
\begin{equation}
    \beta(h) = \lambda_1^\beta \sigma(k(h)) + \lambda_2^\beta \overline{k(h)}^2 + \lambda_3^\beta \frac{\overline{|h-\bar{h}|}}{h_0},
\end{equation}
where $\lambda^\beta_{1,2,3}$ are the weights for edges, slopes, and roughness, $\sigma(\cdot)$ is the standard deviation, $k(h)$ is the slope angles divided by $\frac{\pi}{2}$, and $h_0$ is the designed maximum height. In our implementation, $\lambda^\beta_1=0.3$, $\lambda^\beta_2=0.5$, $\lambda^\beta_3=0.2$.

Typically, for our generator trained in Sec. \ref{sec:generating}, the scores are within $[0.02,0.05]$, and the effect of the noise inputs on the score is empirically less than $0.005$ even for extreme cases. Yet, with our learning technique, the sampled terrains can even reach a high score of $0.08$, or $10^{-3}$-level deviation from a specified score value, while maintaining the diversity.

To demonstrate the efficacy of our method, we designed two tasks for generation: one called "hard" for the very difficult terrains with a high score, and the other called "medium" for terrains with a score close to $0.055$ which is also hard to achieve via random sampling. The performance will be displayed in Sec. \ref{subsec:show_dist}.

The reward function $R(\cdot)$ for the "hard" task is defined as 
\begin{equation}
    R(h)=\epsilon + 10^{150\beta(h)-6},
\end{equation}
which makes the probability of sampling a terrain $\sim 30$ times higher if the score is $0.01$ larger according to (\ref{eq:property}).

The reward function $R(\cdot)$ for the "medium" task is defined as
\begin{equation}
R(h) = \epsilon + \exp(\frac{0.1}{|\beta(h)-0.055|+0.005}-10),
\end{equation}
which forms a steep spike around $0.055$. 

\subsection{Neural networks for flow approximation}

\begin{figure}[t]
  \centering
    \includegraphics[width=86mm]{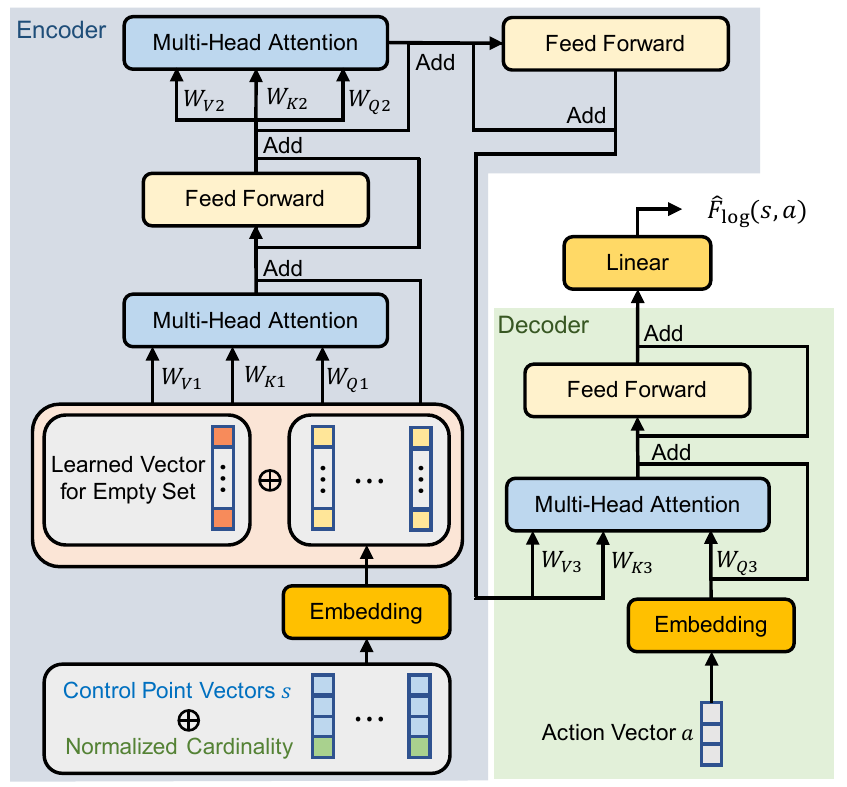}
    \caption{
      The neural network structure for $\hat{F}_{\log}$ approximation.
    } 
    \label{fig:model}
  \vspace{-4mm}
\end{figure}

The neural network structure we used for $\hat{F}_{\log}$ approximation with non-termination is illustrated in Fig. \ref{fig:model}. To tackle the variable cardinality of states, we took an encoder-decoder structure, where the encoder encodes the state-related information, and the decoder couples the state-related information and the action-related information.

\subsubsection{Encoder} 
The 3-d point vectors $p_1,\dots,p_l$ in a state $s$ are first concatenated with a 1-d cardinality-related value to inform the model of the "progress". With the maximum cardinality $60$, we define this normalized cardinality value as $\frac{\text{card}(s)}{0.5\times 60}-1$. These vectors are then embedded into 128-d vectors. To deal with the case of the empty set $s$, we introduce an additional learned 128-d vector which is concatenated to the embedded vectors, making in total $\text{card}(s)+1$ vectors of length $128$ to represent the states.

These vectors are then fed into two repetitive combinations of the multi-head attention module \cite{vaswani2017attention} and the feed-forward layer. The multi-head attention module is with embedding dimension $128$ and the number of heads $4$, and the outputs are added with the inputs as a residual skip connection. The feed-forward layer is a 128-d linear layer activated by the GELU function \cite{hendrycks2016gaussian}, and also takes a skip connection.

\subsubsection{Decoder}
The 3-d action vector $a$ is first embedded to a 128-d vector. Then a multi-head attention module (with the same structure as in the encoder) takes the outputs of the encoder for the key source and the value source, and the query source is the embedded action vector. Then, with a skip connection from the embedded action vector, the outputs of the multi-head attention module are fed into a feed-forward layer with a skip connection. 

Finally, the outputs of the decoder go through a linear layer to make the final output, i.e, to approximate $\hat{F}_{\log}(s,a)$.

\subsection{Other details for implementation}
\subsubsection{Sampling-based outflow estimation}
For each state, there are more than 10000 feasible actions except for those with $60$ points. For each state-action pair, the neural network is used to compute the flow approximation, which makes the training time-consuming. Instead, we estimate the outflow term $\sum_{a \in \mathbb{A}^*(s')\backslash \{a^T\} } \exp \hat{F}_{\log}(s',a) $ in (\ref{eq:optim}), by randomly sampling $2000$ actions and getting the average flow that is then multiplied by the cardinality of $\mathbb{A}^*(s')\backslash \{a^T\}$. This reduces the training time by 80\%.

\subsubsection{Biasing the $\hat{F}_{\log}(s,a)$ model output}

The neural network outputs are around $0$ initially, which makes an initial guess of the ideal flows. Yet, scores of the initially sampled terrains are within $[0.02,0.05]$, which specifies the termination flow. Therefore, if the neural network outputs too large values compared with the termination flow, flow mismatching between non-terminated state transitions will take the lead in the loss, which hinders the learning for better terrains. If the initial sampled rewards get too large, the sampler will tend to terminate early and get stuck in local exploration. Hence, we add a bias to the $\hat{F}_{\log}(s,a)$ output, $0$ for "hard" and $-6$ for "medium", to seek a balance.

\subsubsection{Clamping the flows}

Flows near the root state are exponentially large, which can lead to the floating point overflow when calculating the loss after episodes of training. To prevent the overflow, we clamp the logarithms of the flows, and bias the neural network outputs instead of the rewards. Still, this can affect the learning process after convergence, which is further discussed in Sec. \ref{subsec:sample_eff}.

\subsubsection{Encouraging exploration}
To encourage exploration, the sampler has a probability of 0.05 to sample a random action per step during training. We also sampled $60$ random trajectories with lengths varying from $1$ to $60$ before training, so that the model can have a basic exploration of the trajectory length.

\section{Results and Discussion}

\begin{figure}[t]
  \centering
    \includegraphics[width=86mm]{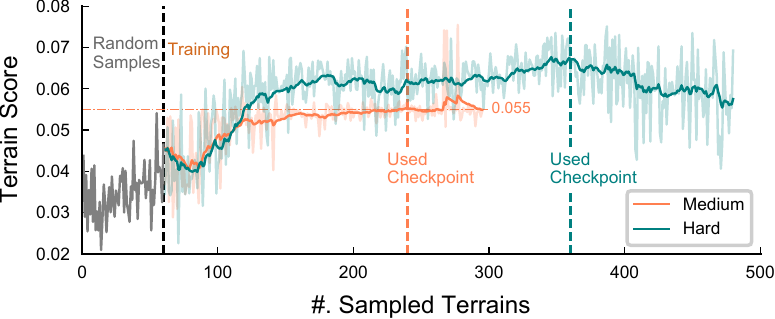}
    \caption{
      Latest sampled terrain scores versus the number of samples during training of the samplers in two tasks. The opaque lines are smoothed values for the original ones in shadow.
    } 
    \label{fig:learning_curve}
  \vspace{-5mm}
\end{figure}

\subsection{Sample efficiency}
\label{subsec:sample_eff}

Fig. \ref{fig:learning_curve} shows how scores of the sampled terrains change during training of the flow approximation models. Both models were well trained within hundreds of samples, exhibiting high data efficiency.

Yet, their performance both went bad after certain iterations, which according to our analyses were due to the clamped flows. The flows from the root state are extremely large and can hardly avoid overflow if we do not use very-high-precision floating-point numbers. After the flows from the root are clamped, further flows become mismatched, which can mislead the optimization. That said, if we do not clamp the flows, the loss will go to "NaN".

\subsection{Distributions of high-quality trajectories}
\label{subsec:show_dist}

\begin{figure}[t]
  \centering
    \includegraphics[width=86mm,trim=2.5mm 2.5mm 1.5mm 2.5mm,clip]{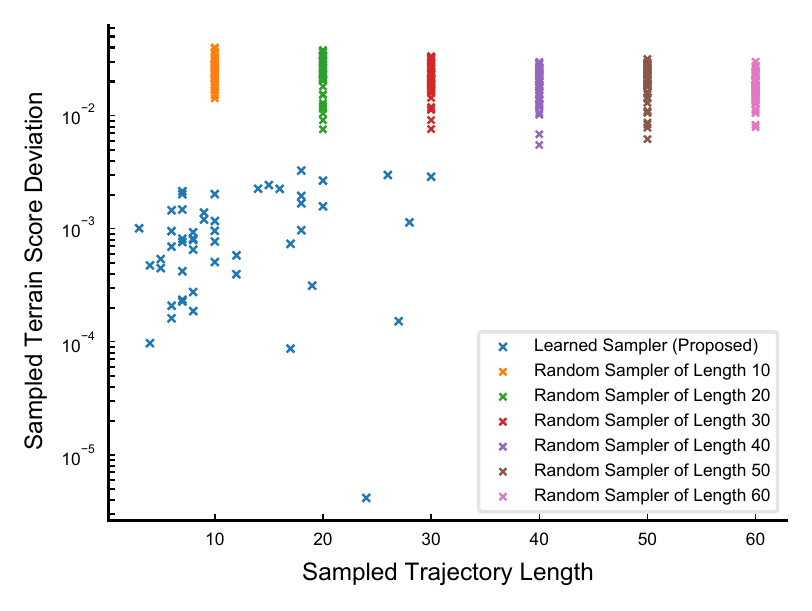}
    \caption{Comparison of the score deviation from $0.055$ for different samplers in the "medium" task, 50 points per sampler. Our method can reach $10^{-3}$-level deviation, which is comparable to the effect of noise.}
    \label{fig:medium_dis}
  \vspace{-0mm}
\end{figure}

\begin{figure}[t]
  \centering
    \includegraphics[width=86mm,trim=2.5mm 2.5mm 1.5mm 2.5mm,clip]{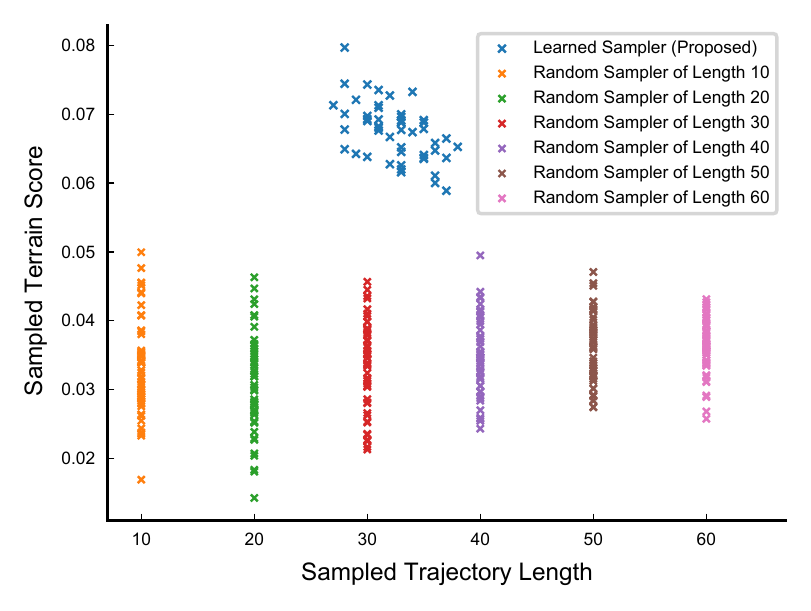}
    \caption{Comparison of the score distributions for different samplers in the "hard" task, 50 points per sampler. Our method can reach much higher scores than the random samplers.}
    \label{fig:hard_dis}
  \vspace{-4mm}
\end{figure}

Fig. \ref{fig:medium_dis} and Fig. \ref{fig:hard_dis} show the length and score (deviation) distributions of different samplers for the "medium" and "hard" tasks respectively. Different trajectory lengths for the random samplers were tested because it is hard to determine an optimal one. 

Besides the outstanding performance, our learned samplers generate trajectories of varying lengths, which also reflects the sample diversity that can be further seen in the dataset.  

\subsection{Policy evaluation}
Extensive evaluation for different robot platforms and policies are done and posted on the codebase website, including the velocity tracking challenge and the fall recovery challenge. Perceptive policies in \cite{rudin2022learning} outperforms non-perceptive policies regarding velocity tracking, and a better initial state distribution in \cite{zhang2022accessibility} and \cite{yang2020multi} can outperform the random distribution regarding fall recovery. Still, the fall recovery policies can suffer from unexpected collisions with the terrain (see our video attachment).

\section{Future Works}
In this paper, we prototyped the generation of a terrain dataset that can be used as a terrain-robustness benchmark for legged locomotion. Extensive policy evaluation are also done and the results are in line with expectations.

Our future works will focus on three aspects:
\begin{enumerate}
    \item How our dataset can be generated in a goal-conditioned way, i.e., to train one sampler for multiple difficulty;
    \item How our dataset can be used in a data-driven way to train terrain-robust locomotion policies;
    \item How our method can be generalized to the generation of non-rigid terrains.
\end{enumerate}

In brief, we hope the extensions of our proposed method can change the way researchers train and evaluate policies.


\bibliographystyle{IEEEtran}
\balance
\bibliography{IEEEabrv,Reference}

\end{document}